\pdfoutput=1
\documentclass[make,preprints,accept,pdftex,moreauthors]{Definitions/mdpi} 
\firstpage{1} 
\pubvolume{1}
\issuenum{1}
\articlenumber{0}
\pubyear{2026}
\copyrightyear{2026}
\datereceived{ } 
\daterevised{ } 
\dateaccepted{ } 
\datepublished{ } 

\Title{Deductive Logic in Language Models: Horizontal vs Vertical Reasoning}

\Author{Davide Maltoni, and Matteo Ferrara}

\AuthorNames{Davide Maltoni, and Matteo Ferrara}

\address{%
Department of Computer Science and Engineering, University of Bologna, Italy; \{name.surname\}@unibo.it}

\abstract{Recent language models exhibit significant logical reasoning abilities, yet the mechanisms supporting deductive inference remain poorly understood. This paper studies small transformer-based language models trained from scratch on multi-step deductive tasks, focusing on the distinction between horizontal reasoning, where intermediate steps are generated autoregressively, and vertical reasoning, where inference unfolds implicitly across layers before the first output token is produced.
We analyze two synthetic tasks: logical consequence over chains of symbolic implications and root-to-leaf navigation in binary trees. Mechanistic interpretability reveals that Chain-of-Thought supervision enables models to learn rule-based inference rather than statistical shortcuts. In the horizontal setting, a shallow attention-only model develops interpretable circuits for rule completion, rule chaining, and final decision making, largely implemented through induction-head-like mechanisms. We further introduce a truncated pseudoinverse method to decode the information carried by queries, keys, and values.
For vertical reasoning, Chain-of-Thought appears to act less as explicit step-by-step guidance and more as a form of curriculum learning, helping the model acquire increasingly complex reasoning patterns. Without Chain-of-Thought, models tend to memorize or exploit dataset biases. These results provide a low-level account of how transformers can implement deductive reasoning and suggest how Chain-of-Thought may serve different functions in horizontal and vertical reasoning.}


\begin{document}


\section{Introduction}

Recent Large Language Models (LLMs) have demonstrated remarkable capabilities in reasoning and problem-solving \citep{Huang2023}. Many approaches have focused on enhancing logical reasoning in LLMs, with a growing body of work introducing formal and symbolic logic-based benchmarks \citep{Liu2025}. While much of the literature emphasizes solving reasoning benchmarks, comparatively less attention has been devoted to understanding and explaining the underlying low-level computational mechanisms. Yet, interpretability is crucial for designing more robust and targeted models, that are less prone to errors. Indeed, although recent models excel in controlled-setting logic tasks, they often struggle to deal with complex problems requiring multiple reasoning steps, Out-of-Distribution (OoD) data, adversarial perturbations, and semantically equivalent variants \citep{Creswell2023,Bowen2024,Li2025,Liu2025}.

Research on explaining logical reasoning in language models can be broadly divided into two main lines of work: (i) analyses of medium-sized pretrained LMs aimed at elucidating the concrete mechanisms they deploy at inference time (\citep{Kim2025,Hong2025,Chen2026}), and (ii) studies that train comparatively small LMs from scratch to examine their learning and generalization capacities, as well as the emergence of low-level mechanisms and circuits (\citep{Guo2025,Ye2025,Brinkmann2024}). The two approaches are both informative and largely complementary, as discussed in Section 2 (related works).
In this paper, we follow the second line of work to investigate the low-level computational mechanisms a transformer-based LM can deploy to solve a multi-ops deductive logic task, one that cannot be addressed through purely statistical cues. To this end, we focus on two reasoning paradigms, which we refer to as horizontal and vertical reasoning.
\begin{itemize}
    \item \textit{Horizontal} reasoning incrementally produces tokens in an autoregressive manner, allowing the model to leverage previously generated tokens for subsequent reasoning steps. This reasoning mode can be effectively elicited through Chain-of-Thought (CoT) training.
    \item \textit{Vertical} reasoning unfolds across the layer hierarchy and cannot directly exploit autoregressive processing, because generating the first token often requires that the problem (or a non-trivial part of it) has already been solved. It is also known as implicit reasoning \citep{Ye2025} since it does not verbalize intermediate steps.
\end{itemize}
Of course, mixed paradigms are more likely to be used in practice by LLMs, also to handle NLP complexity and leverage knowledge acquired during pretraining and not included in the prompt.

The first deductive logic problem considered in this paper involves identifying the chain of rules (if any) connecting a premise to a conclusion. This type of reasoning is fundamental to many problem-solving tasks in science and engineering.
Another related problem we consider, recently introduced in \citep{Brinkmann2024}, consists of identifying the root-to-leaf path when navigating a binary tree. Both the problems require a multi-step reasoning with a length of the reasoning chain, that is the number of rules that must be considered/combined to reach the conclusion, significantly larger than in most of the existing works. 

Our study is conducted using mechanistic interpretability tools \citep{Olah2022,Elhage2021,Nanda2023}, and, to simplify the analysis, we look for the simplest architecture capable of solving the task.
Our low-level explanation leverages several interpretation techniques, including ad-hoc visualization of attentions, decoding residual stream content into vocabulary space, averaging across multiple inputs to identify token-independent attention links, and apply a truncated pseudoinverse to reveal the information extracted by queries, keys, and values. Notably, the use of a truncated pseudoinverse in this context is quite novel and may serve as a useful inspection tool for future interpretability research.

Our findings can be summarized as follows:
\begin{enumerate}
    \item A very simple, non-pretrained model trained with Chain-of-Thought (CoT) prompting can learn the underlying inference rules and generalize to novel examples.
    \item Both horizontal and vertical reasoning can emerge when a model is trained with CoT. Yet, CoT plays distinct roles in the two cases: it supports horizontal reasoning through step-by-step task decomposition, and it facilitates vertical reasoning by inducing a form of Curriculum Learning.
    \item Horizontal reasoning is simpler and can be learned by relatively shallow models with limited training data, whereas vertically unfolding reasoning steps across layers requires deeper models and more intensive training.
    \item Training without CoT is substantially more challenging for complex tasks and, in the absence of exploitable dataset biases, often fails to converge to solutions that generalize.
    \item Induction heads \citep{Olsson2022} generally play a central role in the formation of internal circuits \citep{Cammarata2020} that implement the compositional computation behind the deductive reasoning.
\end{enumerate}

The main contribution of this work is an in-depth, low-level account of the mechanisms and circuits that a small LM, trained from scratch with CoT, develops to perform horizontal reasoning. Classical mechanistic analysis tools have been applied for this purpose, and a novel method based on a truncated pseudoinverse is introduced to disentangle the information carried by queries, keys, and values. 
This contrasts with most previous studies, which focus on higher-level descriptions and usually restrict themselves to offering intuitions about the most complex reasoning steps. A further relevant contribution is to analyze the conditions under which horizontal and vertical reasoning can arise, both in the presence and absence of CoT and Curriculum Learning.

After reviewing the related literature in Section 2, we formally define the two tasks in Section 3 and describe the model architecture in Section 4. Section 5 concentrates on horizontal reasoning and offers a low-level mechanistic account of the strategy acquired by a simple 2-layer model trained on Task 1 (with CoT). By contrast, Section 6 examines vertical reasoning and reveals the mechanism learned by a deeper 6-layer model applied to Task 2 (with CoT). In Section 7, we demonstrate that, without CoT, our models are unable to learn either Task 1 or Task 2, except in situations where generation biases can be leveraged to exploit statistical shortcuts. Section 8 then delves further into the role of CoT, showing that it influences horizontal and vertical learning in distinct ways. Finally, Section 9 provides concluding remarks and suggests avenues for future research. All the experiments described in this paper are fully reproducible using the codebase available on \href{https://github.com/MatteoFerrara/Deductive-Logic-in-Language-Models---Horizontal-vs-Vertical-Reasoning}{GitHub}. 

\section{Related works}

\subsection{Assessing deductive reasoning capabilities of LLMs}

Several benchmarks have been proposed for evaluating LLMs on deductive reasoning tasks, including the recently introduced JustLogic \citep{Chen2025}, ProofWriter/RuleTaker \citep{Tafjord2021,Clark2021} and LogicBench \citep{Parmar2024}. Synthetic generation of examples is widely adopted, as it enables systematic control over benchmark difficulty, ensures correctness, and helps minimize bias. For example, JustLogic \citep{Chen2025} was designed to increase the complexity of the task, reduce the dependence on prior knowledge, and enable detailed error analysis, and Mirage \citep{Li2025} was introduced to make the generation of test sets more flexible and the evaluation more comprehensive.

The deductive reasoning tasks considered in this work (introduced in Section 3) are also synthetically generated, but are intentionally simpler than those in the above benchmarks. The simplification comes from removing the complexity and ambiguity of natural language and representing facts and rules using basic symbols. Such design choices allow us to employ a simple, non-pretrained language model for the task, with emphasis on explainability.

\subsection{Previous work on explaining deductive reasoning in LMs}

Table \ref{table:RelLit} summarizes representative works focusing on the explaination of deductive reasoning in LMs. The first rows (with white background) refer to medium-size pretrained LMs (hereafter category A), while the grayed rows denote small models trained from scratch on deductive reasoning tasks (hereafter category B). Both these lines of work are interesting and somewhat complementary. 

In category A studies, the pretrained models are not tuned on the considered deductive reasoning task and often reach an accuracy of just 70-80\%. However, the inference analysis of a sufficient number of properly solved test cases convinced the researchers that the underlying models perform a structured processing decomposable into phases, not an opaque flow of information. Mechanistic tools allowed to describe the high-level reasoning path and information flow and to isolate the model's units (e.g., heads, layers, columns, etc.) carrying out the corresponding steps. Unfortunately, to make the analysis tractable, Category A works typically deal with simple deduction problems requiring the application of a few rules (e.g., a transitive property with two rules and a single shared variable). Furthermore, they do not reveal detailed low-level circuits and computational mechanisms (e.g., how a logical rule is encoded, or how a variable is bound to true or false) and often cannot definitively rule out that the discovered patterns reflect sophisticated statistical shortcuts that merely approximate genuine logical algorithms.

Category B studies training smaller LMs on specific deduction tasks typically achieve near-perfect accuracy, and the architectural simplicity allows to go deeper with the mechanistic analysis. However, demonstrating that a small transformer has the computational capacity to solve a non-trivial logical problem does not imply that larger LMs (including pretrained SOTA models) rely on the same mechanisms to perform logical reasoning.

The approach followed in this paper belongs to category B, since our main aim is demonstrating that a transformer can learn a multi-ops deductive reasoning task in terms of explicit and fully explainable logic steps. The work by Brinkmann et al. \citep{Brinkmann2024} is the most closely related to our study, and its problem formulation, experimental methodology, and results were highly influential in motivating us to explore the differences between horizontal and vertical reasoning. Indeed, although our initial investigations on horizontal reasoning were conducted \citep{Maltoni2025} in parallel with \citep{Brinkmann2024}, we later became aware of that work, which enabled us to broaden our analysis and gain new insights.

\begin{table*}[t]
\caption{Recent representative works on explanation of logic reasoning in LM. Rows with white background refer to studies on medium-size pretrained models, while grayed rows denote studies on smaller models trained from scratch. The task complexity is characterized by the length of the reasoning chain (Ops) and the use of Natural Language (NL) in the prompt instead of simple symbols.}
\label{table:RelLit}

\scriptsize
\renewcommand{\arraystretch}{1.15}
\renewcommand{\tabularxcolumn}[1]{m{#1}}
\setlength{\tabcolsep}{3pt}

\begin{tabularx}{\textwidth}{
    @{}
    >{\RaggedRight\arraybackslash\hspace{0pt}}m{1.25cm}
    >{\RaggedRight\arraybackslash\hspace{0pt}}X
    >{\RaggedRight\arraybackslash\hspace{0pt}}m{2.2cm}
    >{\RaggedRight\arraybackslash\hspace{0pt}}m{2.45cm}
    >{\RaggedRight\arraybackslash\hspace{0pt}}X
    @{}
}
\toprule
\textbf{Work} & \textbf{Title} & \textbf{Model(s)} & \textbf{Task Complexity} & \textbf{Main findings} \\
\midrule

Kim 2025 \citep{Kim2025} &
Reasoning Circuits in Language Models: A Mechanistic Interpretation of Syllogistic Inference &
GPT-2, Pythia, Qwen, Llama.\newline
Train: pretrained &
Ops: 2 (transitive property)\newline
NL: moderate &
High-level circuit descriptions, shared among different models. Contamination by pre-training knowledge. \\
\midrule

Hong 2025 \citep{Hong2025} &
A Implies B: Circuit Analysis in LLMs for Propositional Logical Reasoning &
Mistral and Gemma (up to 27B)\newline
Train: pretrained &
Ops: 2 (2 rules, 5 variables)\newline
NL: moderate &
Identifies high-level sub-circuits for distinct steps such as context retrieval, information processing, and answer generation. \\
\midrule

Chen 2026 \citep{Chen2026} &
Towards a Mechanistic Understanding of Propositional Logical Reasoning in Large Language Models &
Qwen3 (8B and 14B)\newline
Train: pretrained &
Ops: 1 and 2 (11 boolean algebra rules)\newline
NL: no &
Focus on computational patterns rather than circuits, as they generalize better. Identification of high-level mechanisms generalizing across architectures. \\
\midrule

\rowcolor{gray!15}
Guo 2025 \citep{Guo2025} &
How Do LLMs Perform Two-Hop Reasoning in Context? &
Small (3 layers)\newline
transformer\newline
Train: from scratch &
Ops: 2 (reasoning chains with distractors)\newline
NL: no &
Medium-size pretrained LLMs struggle with distractors unless fine-tuned. A small 3-layer model can solve the task via simple, interpretable mechanisms. \\
\midrule

\rowcolor{gray!15}
Ye 2025 \citep{Ye2025} &
How do Transformers Learn Implicit Reasoning? &
Small (3 layers)\newline
transformer\newline
Train: from scratch &
Ops: 2 (traversing an implicit entity)\newline
NL: no &
Analysis of generalization to out-of-distribution data and learning phases. Latent-space representation clustering correlates with generalization. \\
\midrule

\rowcolor{gray!15}
Brinkmann 2024 \citep{Brinkmann2024} &
A Mechanistic Analysis of a Transformer Trained on a Symbolic Multi-Step Reasoning Task &
Small (6 layers)\newline
transformer\newline
Train: from scratch &
Ops: up to 15 (root-to-leaf traversal in a binary tree)\newline
NL: no &
The task is solved by climbing one tree level per layer. For deeper trees, parallel paths are followed and subsequently fused. \\

\bottomrule
\end{tabularx}
\end{table*}

\subsection{Interpretability techniques}

Numerous techniques have been proposed to study the internal approach used by a Transformer-based LM to solve a task \citep{Ferrando2024}. While intrinsic methods focus on the training process, post-hoc techniques are applied at inference-time. Post-hoc investigations can be passive (e.g., analyzing attention patterns) or active (e.g., injecting or replacing specific activations). 

A common strategy to study the computation performed by a model is tracing the flow of information through its most contributing components. To this purpose: (i) attribution methods \citep{Madsen2022} estimate the contribution of individual elements (e.g., specific input tokens); and (ii) causal mediation analysis \citep{Vig2020} seeks causal relationships between model components and predictions and is often carried out through activation patching \citep{Heimersheim2024}. 

Decoding internal representations is another important aspect of interpretability. Probing \citep{Belinkov2022} is a common technique where a simple model (the probe) is trained on top of LM embeddings to predict a given property; however, probing is often criticized for revealing correlations instead of causations. More recently, Sparse AutoEncoders (SAEs) have been used to disentangle features in superposition, thus extracting monosemantic and more interpretable features \citep{Sharkey2022}. Another useful approach is decoding into vocabulary space (see LogitLens and related methods in \citep{Ferrando2024}), where internal representations are searched for the presence of specific tokens.

Some researchers argue that discovering low-level circuits \citep{Cammarata2020} implementing atomic functions is necessary for fully reverse-engineering the algorithm learned by a model. Circuits can be viewed as subgraphs of the entire model computation graph. Mechanistic interpretability \citep{Olah2022,Elhage2021}, often combined with causal mediation analysis, provides tools to support such low-level circuit discovery. Several studies have successfully applied these techniques to explain how LMs solve nontrivial tasks  \citep{Nanda2023,Stolfo2023,Wang2023,Heimersheim2023,Hanna2023}. 

In this paper, we use post-hoc, passive inspection techniques typical of mechanistic interpretability to identify and explain the low-level Transformer circuits solving the considered deductive reasoning task. For residual stream decoding into vocabulary space, we rely on the well-known LogitLens approach. Additionally, we introduce a novel technique based on truncated pseudoinverse to decode the information carried by queries, keys, and values into token space.

\subsection{Induction heads}

Induction heads were first identified in \cite{Elhage2021} and later studied in \cite{Olsson2022}, who highlighted their important role in in-context learning. Induction heads implement a pattern-matching approach capable of completing a sequence [A], [B] … [A] with the token [B]. In other words, they search the previous tokens in the sequence for a pair of the form [A], [B] and, in case of success, [B] is released for sequence completion. The implementation of this mechanism requires at least two layers (even without MLPs) because the model must first copy the key (i.e., [A]) into the same column of the value to disclose ([B]). The authors of \cite{Nanda2023} argued that a phase-change occurs during training when induction heads are formed, and the model shifts from memorizing patterns to generalizing.

In the paper, we show that induction heads play a crucial role in implementing the logic inference underlying our deductive reasoning task. 

\section{Task design} \label{sec:TaskDesign}

\subsection{Task 1: Logical Consequence} \label{sec:TheTask}

We propose a propositional logic inference task where a set of five implications (i.e., Horn clauses with a single literal in the head) is given as a set of true rules, along with an additional implication serving as query. The goal is to decide whether the query is a logical consequence of the five true rules. Formally, the task can be expressed as follows:

\begin{center}
$a_1 \rightarrow b_1,a_2 \rightarrow b_2, \ldots, a_5 \rightarrow b_5 \models q_0 \rightarrow q_1$
\end{center}

\noindent
where $a_i \rightarrow b_i$ for $i=1,\ldots,5$ are the five true implications, and $q_0 \rightarrow q_1$ is the query.

Instances of the problem are generated by substituting $a_i$, $b_i$, $q_0$ and $q_1$ with literals sampled from the first 20 uppercase letters of the English alphabet. For example:

\begin{center}
$K \rightarrow F,C \rightarrow D,B \rightarrow C,A \rightarrow B,D \rightarrow E \models A \rightarrow E$
\end{center}

\noindent
is a positive example since $A \rightarrow E$ is a logical consequence of the five given rules (i.e., a 4-ops reasoning chain can be found in the true implications, starting with $A$ and ending with $E$), while:

\begin{center}
$E \rightarrow F,C \rightarrow K,B \rightarrow C,A \rightarrow B,D \rightarrow E \models A \rightarrow F$
\end{center}

\noindent
is a negative example since the chain is broken at $K$ after 3 steps.

To simplify logic inference, the implications are generated without cycles, and for positive examples, there is exactly one path connecting $q_0$ to $q_1$. Details of the dataset generation process are provided in Appendix \ref{sec:AppendixA}. It is worth noting that: (i) for both positive and negative instances, the reasoning chain to reconstruct has variable length (from 1 to 5); (ii) the resulting generation space is huge: about 469 billion of different (negative or positive) examples. Therefore, despite its apparent simplicity, the task cannot be solved through pure memorization or by smooth interpolation over a small training dataset.

In the non-CoT version of the task, the model output is a single token: 1 when the query is a logical consequence of the given rules and 0 otherwise. 

To help the LM solve the task, instead of providing only binary supervision, we can use Chain-of-Thought (CoT) during training \citep{Wei2022}. Hence, for the two examples above, the corresponding outputs are:

\begin{center}
$A \rightarrow B,B \rightarrow C,C \rightarrow D,D \rightarrow E,\_ \rightarrow \_ - 1$
\end{center}
\begin{center}
$A \rightarrow B,B \rightarrow C,C \rightarrow K,\_ \rightarrow \_,\_ \rightarrow \_ - 0$
\end{center}

\noindent
where: (i) the implications are ordered according to the reasoning chain leading from $q_0$ to $q_1$ (or to the last literal before a chain break); (ii) the underscore character $'\_'$ is used as padding to complete a 5-ops chain\footnote{While a variable-length output using an End-of-Sequence (EoS) token would be a more elegant solution, we opted for padding to simplify the application of some visualization and explanation techniques (discussed later).}, and (iii) the last character represents the final decision.

\subsection{Task 2: R2L navigation in a binary tree}\label{sec:task2}
\begin{figure}[ht]
\centering
\includegraphics[scale=0.21]{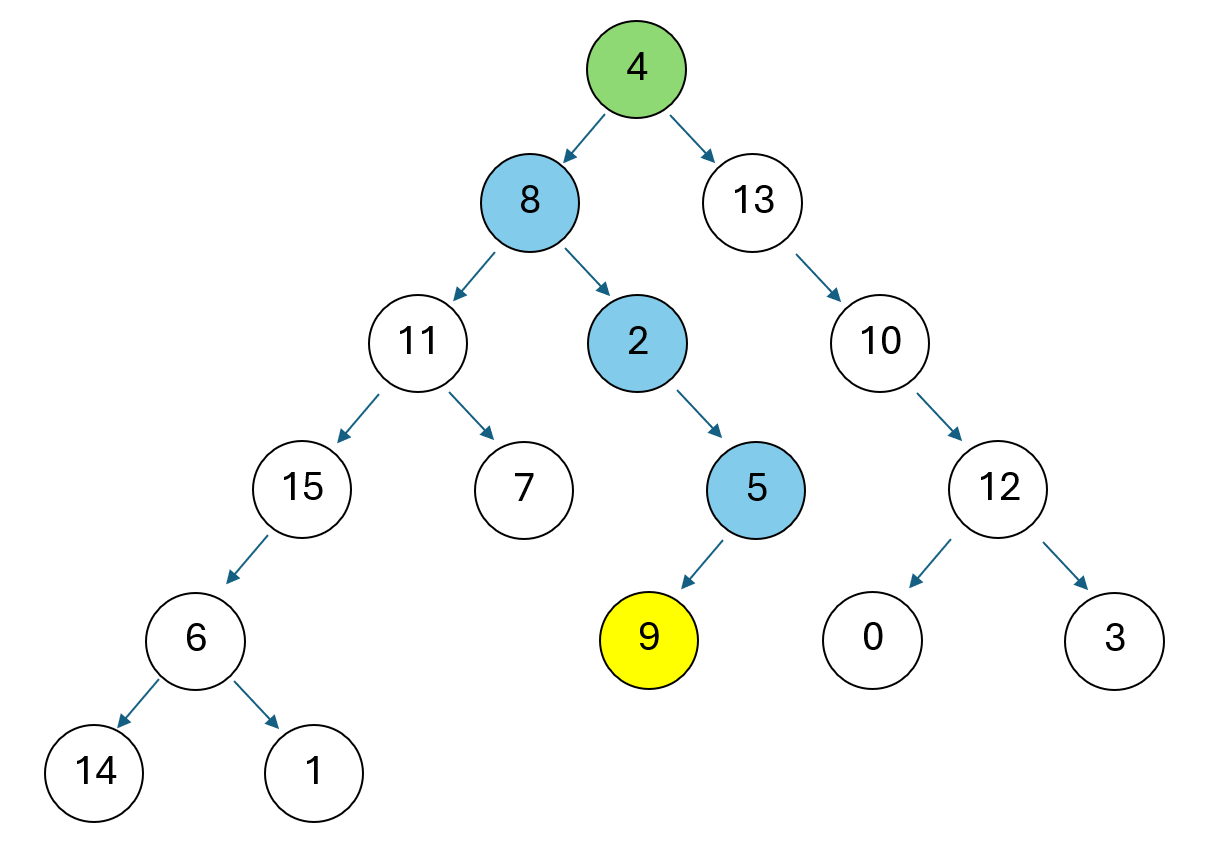}
\caption{An example of a binary tree with 16 nodes. Given the root (green) and the target leaf (yellow), the task consists of generating the root-to-leaf path: 4 8 2 5 9. The prompt includes the (randomly shuffled) link list: 11→7, 8→2, 12→3, 11→15, 6→14, 4→8, 10→12, 15→6, 2→5, 4→13, 12→0, 6→1, 8→11, 5→9, 13→10 followed by the query | 9 : 4.}
\label{figure:TreeExample}
\end{figure}

This task, introduced in \citep{Brinkmann2024}, considers binary trees with 16 nodes, and given a leaf node, requires producing the sequence of traversed nodes when navigating from the root to the leaf (an example is shown in Figure \ref{figure:TreeExample}). The tree is given in the prompt as an unordered list of parent-child node pairs, hence similarly to Task 1, solving the problem can be viewed as navigating a chain of connections. However, there are some key differences in the two tasks:
\begin{itemize}
\item while moving up in a tree (leaf to root) is straightforward since each node has one parent, the contrary (root to leaf navigation) requires to select the right child for nodes with two children (nodes 4 and 8 in the example of Figure \ref{figure:TreeExample}).
\item in a tree with 16 nodes, if the tree is highly unbalanced the reasoning chain can be up to 15-ops.
\item the leaf node given as the query always exists, so unlike in Task 1, here there is always a root-to-leaf (unbroken) path.
\end{itemize}

Also in this task the generation space is huge: in fact, according to Appendix B.2 of \citep{Brinkmann2024} the number of labeled binary trees with 16 nodes is about $2\times10^{19}$. 

During training, supervision is provided by specifying the ordered list of nodes in the expected root-to-leaf path. Even if this can be viewed as CoT supervision, horizontal reasoning is not straightforward here because of the binary choices required at some nodes (more in Section \ref{sec:Vert}).

As a further variation of this task, we introduce a non-CoT setting in which the expected output is limited to a single token: 0 or 1. Here, 0 indicates that the leaf node is not reachable from the root, while 1 indicates that it is reachable. To achieve this, the generated trees are adjusted by removing a single link either on the true path (yielding output 0) or on an alternative branch (yielding output 1).

\section{The model architecture}\label{sec:ModArc}

All experiments in this work employed a GPT-2-style architecture trained from scratch (i.e., without pretraining). A straightforward implementation of this decoder-only model, referred to as NanoGPT, was introduced in \cite{Karpathy2022} and is also provided in the TransformerLens framework \citep{Nanda2022}. The base model consists of $l$ = 6 layers, each containing $h$ = 8 attention heads, and an MLP with Gelu activation (if $MLP_{presence}=True$). The residual stream has dimensionality $d_{model} = 64$. Notably, the same model architecture was also adopted in \citep{Brinkmann2024}. In some experiments, the base model configuration $(l,h,d_{model},MLP_{presence})$ was simplified to make mechanistic interpretability more tractable (further details are given in successive Sections, and a complete overview of model parametrization in the different experiments is provided in Appendix \ref{sec:AppendixC}). 

When the trained LM is used in inference mode, we always pick the most probable token from the logits (i.e., greedy decoding) and produce the output token by token in autoregressive mode. Model accuracy is measured using full-sequence accuracy, meaning that all generated output tokens must be correct.

\section{Horizontal Reasoning: Learning Task 1 with CoT}\label{sec:HorReas}

In this Section, we demonstrate that training a simplified 2-layer model with CoT on Task 1, even with a small training set, leads to the emergence of a fully explainable horizontal reasoning process.

\subsection{Training}

A simplified model architecture with $(l=2,h=1,d_{model}=128,MLP_{presence}=False)$ was powerful enough to solve Task 1 with CoT training. Using a single attention head ($h:8\rightarrow1$) simplified interpretability but required doubling $d_{model}:64\rightarrow128$.
We generated a dataset of 4,096 examples (50\% positive and 50\% negative), which was randomly split into a training set (3,072 examples, 75\%) and a validation set (1,024 examples, 25\%). Despite the small size of the training set (compared to the vast input space), the LM learns the task (accuracy is \textasciitilde100\%) in just a few hundred epochs. It also perfectly generalizes on the validation set. Refer to Appendix \ref{sec:AppendixC} for further training details. When repeating the experiment with different training sets and different random initial weights for the LM, we observed that approximately 20\% of the runs converged within 250 epochs. The convergence rate increased to 80\% when using the full NanoGPT architecture. However, for the purposes of this study, the success rate across runs is not critical — our goal is to analyze the inference behavior of a few models that have reached convergence.

\subsection{Mechanistic explanation}\label{sec:MechExpl}

The positive prompt C>D,A>B,B>C,E>F,D>E|A>F (expected output: A>B,B>C,C>D,D>E, E>F-1) is used as a guiding example to explain the inference details of a trained LM that has reached convergence. In this case, a complete chain (5-ops) must be reconstructed to derive F from A. After presenting the tool and interpretability techniques used for our analysis, we show that the inference process can be decomposed into three stages: (i) Rule completion, (ii) Rule chaining, and (iii) Start and final decision. Finally, the insights obtained from the guiding example are confirmed using aggregated statistics computed over a larger set of validation examples (Subsection \ref{subsec:FurtherExp}).

\subsection{Tool and investigation techniques}

\begin{figure*}[ht]
\centering
\includegraphics[scale=0.4]{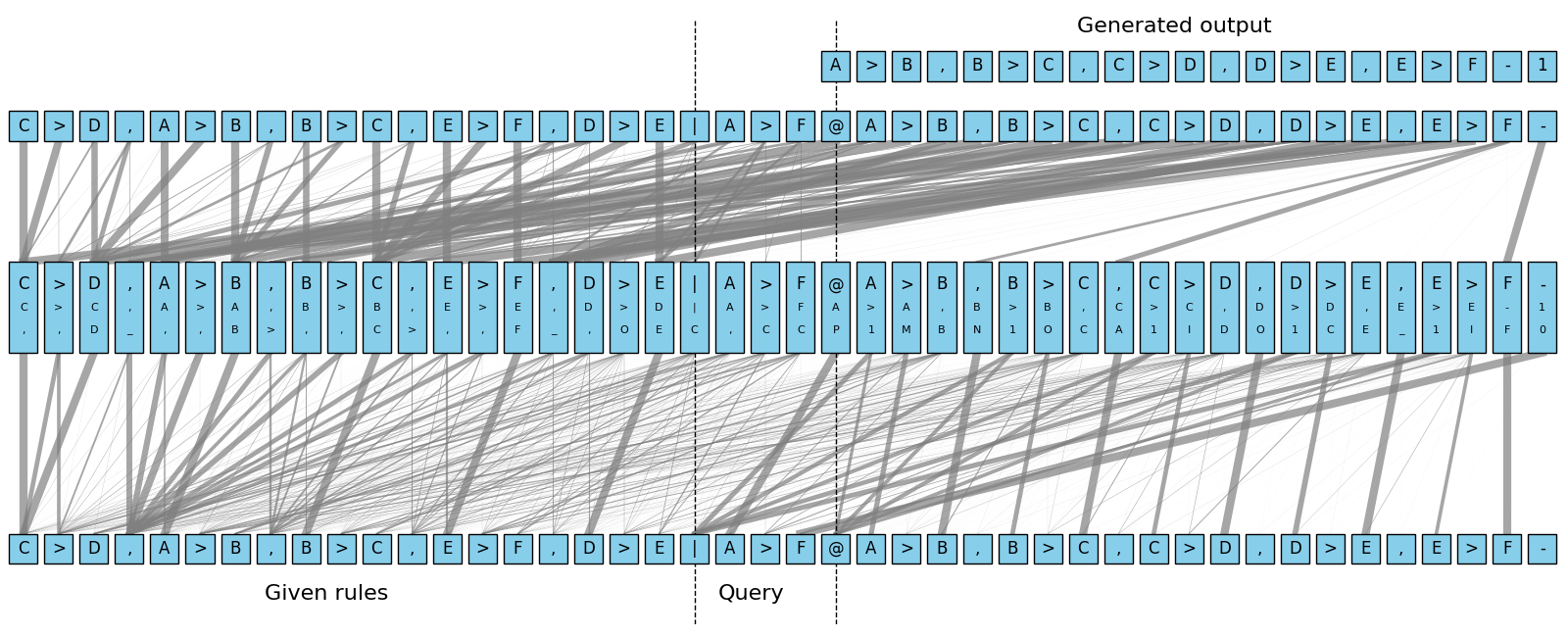}
\caption{An example of output produced by the developed visualization tool. The explanation is in the main text.}
\label{figure:VisToolOutputExample}
\end{figure*}

Figure \ref{figure:VisToolOutputExample} shows the output of the visualization tool developed for this study. To facilitate access and inspection of internal data, the tool was built on top of the TransformerLens framework \citep{Nanda2022}. The bottom row corresponds to the input and includes: (i) the prompt (on the left) divided into the Given rules and Query parts, and (ii) the shifted generated output (on the right) produced by the autoregressive generation process (compare with the generated output in the top row). The second row from the bottom represents the first layer. The character visualized at the top of each box is a copy of the input token, included only for visual alignment. The second and third characters (in smaller font) correspond to the top and second-ranked tokens decoded from the residual stream after processing by Layer 1. This decoding is based on the LogitLens approach \citep{Nostalgebraist2020}, where the final LM head is applied to the residual stream to identify the tokens with the highest logit values. This approach has several limitations because the residual stream internal representation can change along the layers, but in our simplified architecture it is effective because: (i) the encoder and decoder share weights, and (ii) the internal layer is close to both the input and output. The third row from the bottom corresponds to the second layer, and the last partial row shows the generated output. Links between input and Layer 1, and between Layers 1 and 2, represent attentions, with thickness proportional to attention strength, computed by the self-attention mechanism as:

\begin{center}
$softmax(mask(\frac{QK^T}{\sqrt{d_{head}}}))$
\end{center}

\noindent
where, $Q$ and $K$ denote the query and keys, respectively; $d_{head}=d_{model}$ when operating with a single attention head, and $mask$ refers to the causal masking restricting the model from looking at future tokens.

Displaying all attention links simultaneously makes it difficult to identify interesting circuits. To this purpose, the tool provides two filtering options: (i) applying a threshold to hide low-strength attention links, and (ii) further restricting links between Layers 1 and 2 to those reaching one or more specific output positions (as will be illustrated later). For clarity, residual stream connections (i.e., vertical links across layers) have been intentionally omitted from the visualizations.

Another useful technique to isolate specific subsets of attention links is computing their strength according to the average strengths over a dataset of prompts. The dataset can be the entire training set or a subset, such as only positive or only negative examples. Since averaging removes token-dependent attention patterns, the remaining links are those depending only on positional information (introduced by positional encodings).

Finally, to better understand the information conveyed by each attention link, we introduce a technique based on truncated pseudoinverse to decode the content of the Query, Key and Value associated with each link into vocabulary space. It is well known that the matrices $W_Q$, $W_K$ and $W_V$ extract information from the residual stream by projecting it into subspaces. However, the semantics of these projected vectors differ from those in the residual stream, making direct decoding with the final decoder ineffective. Therefore, we compute a truncated pseudoinverse (using truncated SVD) to project these subspace vectors back into the residual stream space, where the LogitLens approach can then be used for decoding (see Appendix \ref{sec:AppendixB} for more details). Truncated SVD retains only the dimensions associated with the largest singular values, discarding low-relevance components. In our setting, this allows us to preserve the most relevant information extracted by the Query, Key and Value matrices while filtering out other information in the residual stream. A threshold parameter, defined as the cumulative sum of top singular values preserved, controls the amount of information (i.e., subspace dimensions) to retain. In the next subsection, examples are provided.

\subsection{Rule completion}\label{sec:rulecompl}

\begin{figure*}[ht]
\centering
\includegraphics[scale=0.8]{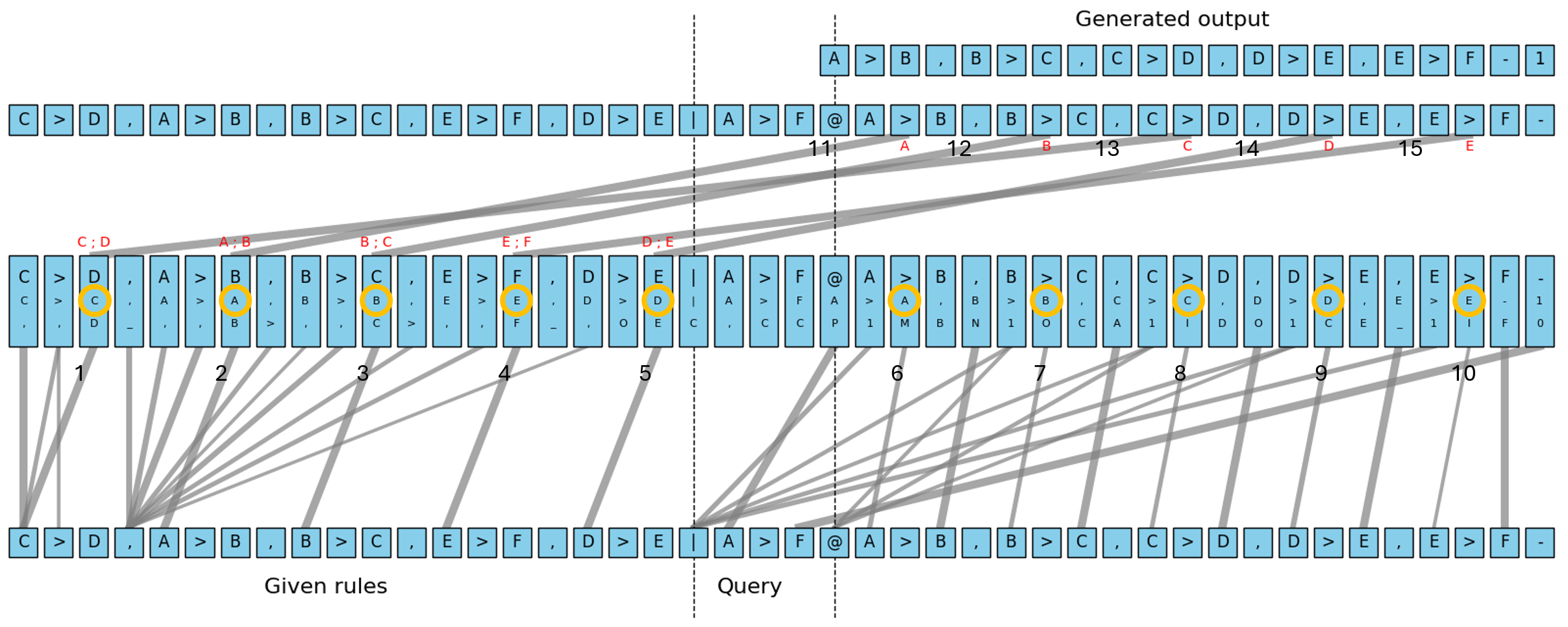}
\caption{Circuits involved in Rule completion. The explanation is in the main text.}
\label{figure:CircuitsRuleCompletions}
\end{figure*}

Figure \ref{figure:CircuitsRuleCompletions} shows the output of the visualization tool when the threshold is set to 0.4 and only attention links from Layer 1 to Layer 2 reaching the $'>'$ tokens in the output are displayed. The three red letters associated with each link represent the decoded content of the Query (top), Key (bottom–left), and Value (bottom–right), computed using the truncated pseudoinverse technique described in the previous subsection and in Appendix \ref{sec:AppendixB}. Obviously, the Query is extracted from the residual stream at the target position, while Key and Value are extracted from the residual stream at the source position. 

Links 1-10 are token-independent attention links (as verified using the averaging technique). These links copy their source token: links 1-5 copy two steps forward, while links 6-10 copy one step forward. This behavior can be confirmed by looking at the top-1 token decoded from the Layer 1 residual stream at the target positions (highlighted with orange circles in the figure). Links 11-15 are responsible for rule completion. Let’s consider link 11 (12-15 work in the same way): it extracts A as the query from the target residual stream and searches for an implication of the form A>X to retrieve the symbol X that completes the rule. The query-key match succeeds because a valid\footnote{The concept of valid position is discussed in more detail in the next subsection.} source position contains A as the key. The released value B is then copied forward and used to generate the correct output token. This copy-search-retrieve behavior corresponds to the well-known induction head mechanism \citep{Olsson2022}.

\subsection{Rule chaining}\label{subsec:RuleChaining}

\begin{figure*}[ht]
\centering
\includegraphics[scale=0.8]{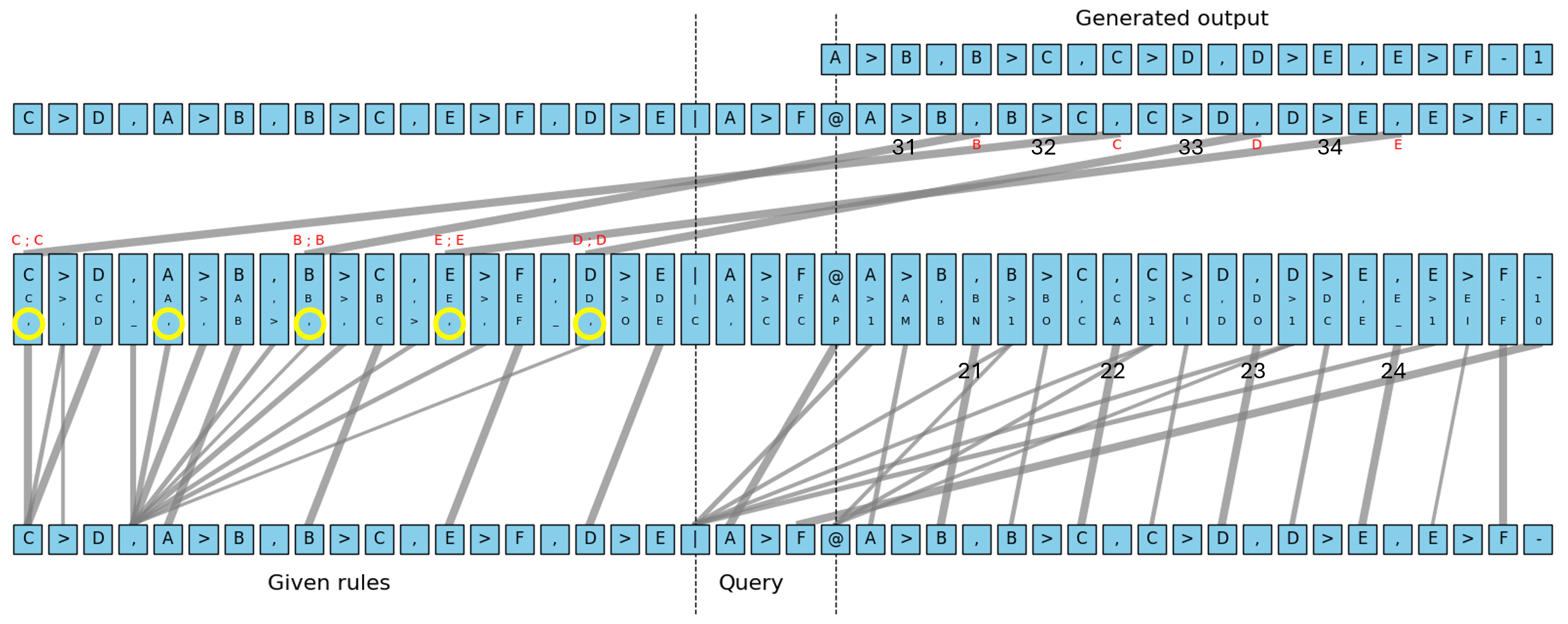}
\caption{Circuits involved in Rule chaining. The explanation is in the main text.}
\label{figure:CircuitsRuleChaining}
\end{figure*}

Figure \ref{figure:CircuitsRuleChaining} shows the visualization tool output when the threshold is set to 0.4 and only the attention links from Layer 1 to Layer 2 reaching the $','$ tokens in the output are displayed. Links 21-24 correspond to token-independent attention links (as confirmed by the averaging technique) which copy their source token one step forward; this behavior can be verified by looking at the top-1 token decoded from the Layer 1 residual stream. Links 31-34 are responsible for rule chaining. Let's consider link 31 (32-34 operate similarly): it extracts B as the query from the target residual stream and searches for an implication of the form B>X to connect the tail B of the previous implication (copied forward by link 21) with the head of a new one in the set of given rules. The query-key match succeeds because a valid source position contains B as the key. The released value in this case is still B, which is copied forward and used to generate the correct output token. This process again reflects an induction-head-like mechanism. It is worth noting that when searching for B in the given rules part of the prompt, only positions where B is in the head of a rule are valid. In our syntax-rigid setting, this is simple to enforce because the rule heads are at fixed positions, allowing positional encoding to tag valid search positions. Interestingly, in the positions corresponding to rule heads the second-ranked token decoded from the residual stream is $','$ (highlighted in yellow). This suggests that when Link 31 searches for a B, the query-key matching may consider not only the literal B but also the presence of the $','$ symbol, which appears in the residual streams at both sides of the link.

\subsection{Start and final decision}\label{subsec:StartFinalDec}

\begin{figure*}[ht]
\centering
\includegraphics[scale=0.8]{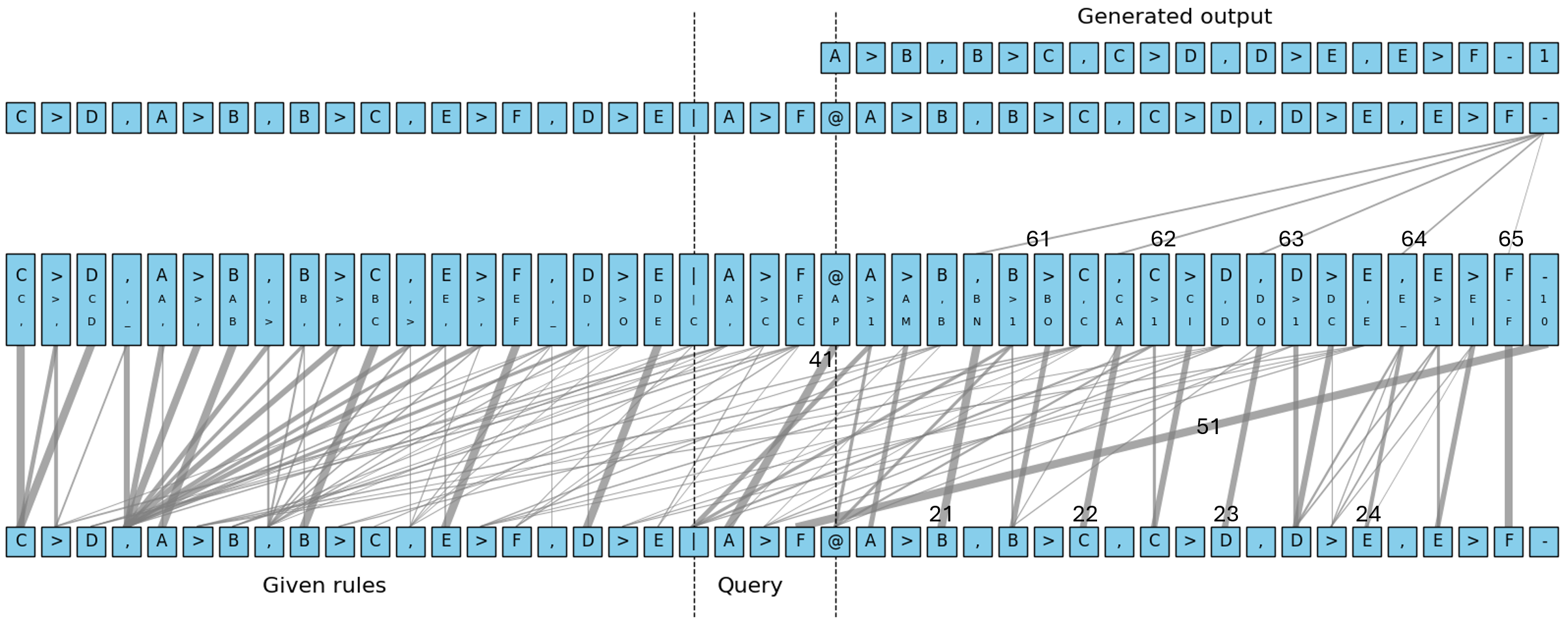}
\caption{Circuits involved in Start and final decision. The explanation is in the main text.}
\label{figure:CircuitsStartFinalDecision}
\end{figure*}

Figure \ref{figure:CircuitsStartFinalDecision} shows the visualization tool output when the threshold is lowered to 0.1 and includes only attention links from Layer 1 to Layer 2 reaching the final $'-'$ token in the output. In this figure, links are visualized based on their average attention strengths over the positive portion of the training set; therefore, all links are token-independent. The beginning of the sequence is straightforward: link 41 copies the head of the query (A in this example) into the residual stream of the start-of-sequence token $'@'$. For the final decision, link 51 copies the tail of the query (F in this example) into the residual stream of the last token $'-'$: Although F does not appear among the top-2 decoded tokens from the residual stream, an extra manual inspection confirmed that it takes the third place in the ranking. Links 61-65 then verify whether F is present in any of the positions corresponding to the tails of implications in the generated chain. This check is necessary because, in positive examples, the chain length can vary between 1 and 5. While the searched position is the expected one for the last implication, in the four preceding cases the positions are shifted one-step forward; however, this does not pose a problem thanks to the positional copies performed by links 21-24, already introduced in the previous subsection.

\subsection{Statistical validation}\label{subsec:FurtherExp}

The mechanistic explanations introduced in Sections \ref{sec:MechExpl} focus on a specific example, but we verified that the same behavior occurs across other positive and negative examples. We also examined additional models (among those reaching convergence) and found that, aside from minor variations, the inference process is largely consistent. 
Here after, we provide further statistical evidence (aggregated over a newly generated datasets of 4096 examples), to confirm the main findings of the rule completion and rule chaining steps. 
\begin{itemize}
\item First, the copy behavior (see links 1..10 in Figure \ref{figure:CircuitsRuleCompletions} and links 21..24 in Figure \ref{figure:CircuitsRuleChaining}) can be confirmed by inspecting with the LogitLens approach the residual stream at the target positions. We found that the average ranking position of the expected token (the one being copied forward) is 1.072 and 1.011 for rule completion and rule chaining, respectively (i.e., the expected token is nearly always the top-1). 

\item A second experiment focused on the search/retrieval capabilities of the induction heads (see links 11..15 in Figure \ref{figure:CircuitsRuleCompletions} and links 31..34 in Figure \ref{figure:CircuitsRuleChaining}): for each target position, we computed the percentage of attention associated with the single link coming from the expected source token, whose position depends on the prompt. The average values obtained are 99.2\% and 99.8\% for rule completion and rule chaining, respectively. 
 
\item In a final experiments, still focused on the search/retrieval capabilities of the induction heads (the same links as in the second experiment), we decoded, with the truncated pseudoinverse approach, the information carried by the Queries, Keys and Values. We found that the average ranking position of the expected tokens is 1.006 for Q, 1.040 for K and 1.000 for V in the case of rule completion and 1.000, 1.015, 1.000 in the case of rule chaining.
\end{itemize}

\section{Vertical Reasoning: Learning Task 2 with CoT}\label{sec:Vert}

To solve Task 2, we initially attempted to reproduce the experiments carried out in \citep{Brinkmann2024}. For this purpose, we started from the code base (and dataset) made available by the authors \citep{Brinkmann2024git}. The same model parameterization was also adopted $(l=6,h=1,d_{model}=128,MLP_{presence}=True)$. Before running the experiments, we found and fixed three issues: 
\begin{enumerate}
    \item in the authors' code, accuracy is computed at the token level instead of at the sequence level (a more stringent requirement to conclude that an example is fully solved). 
    \item a generation bias was identified and removed (see Appendix \ref{sec:AppendixD} for more details). Such a bias was included some statistically exploitable features in the generated trees. It is worth noting that the bias does not invalidate the findings of \citep{Brinkmann2024}, which were confirmed in our experiments. However, as pointed out in Section \ref{subsec:Task2NoCot}, the generation bias has a strong influence on the experiments with no CoT.
    \item to make the tree input representation more explicit, the authors of \citep{Brinkmann2024} used different tokens for the same node depending on its role: parent or child (e.g., '6' and '→6' are different tokens corresponding to node 6 being parent or child in a link). We found such a peculiar encoding to be unnecessary (since the model can easily infer the node role according to the relative position in a token pair) and, for uniformity with Task 1, we removed it.
\end{enumerate}

The model was trained on a dataset of 150K examples, reaching convergence (99.9\% accuracy) in about 50 epochs on both the training and the validation partition (see Figure \ref{figure:TrainGraphTask2} in Appendix \ref{sec:AppendixC}). This is in line with the results in \citep{Brinkmann2024}. 

\begin{figure*}[ht]
\centering
\hspace*{0cm}
\includegraphics[scale=0.28]{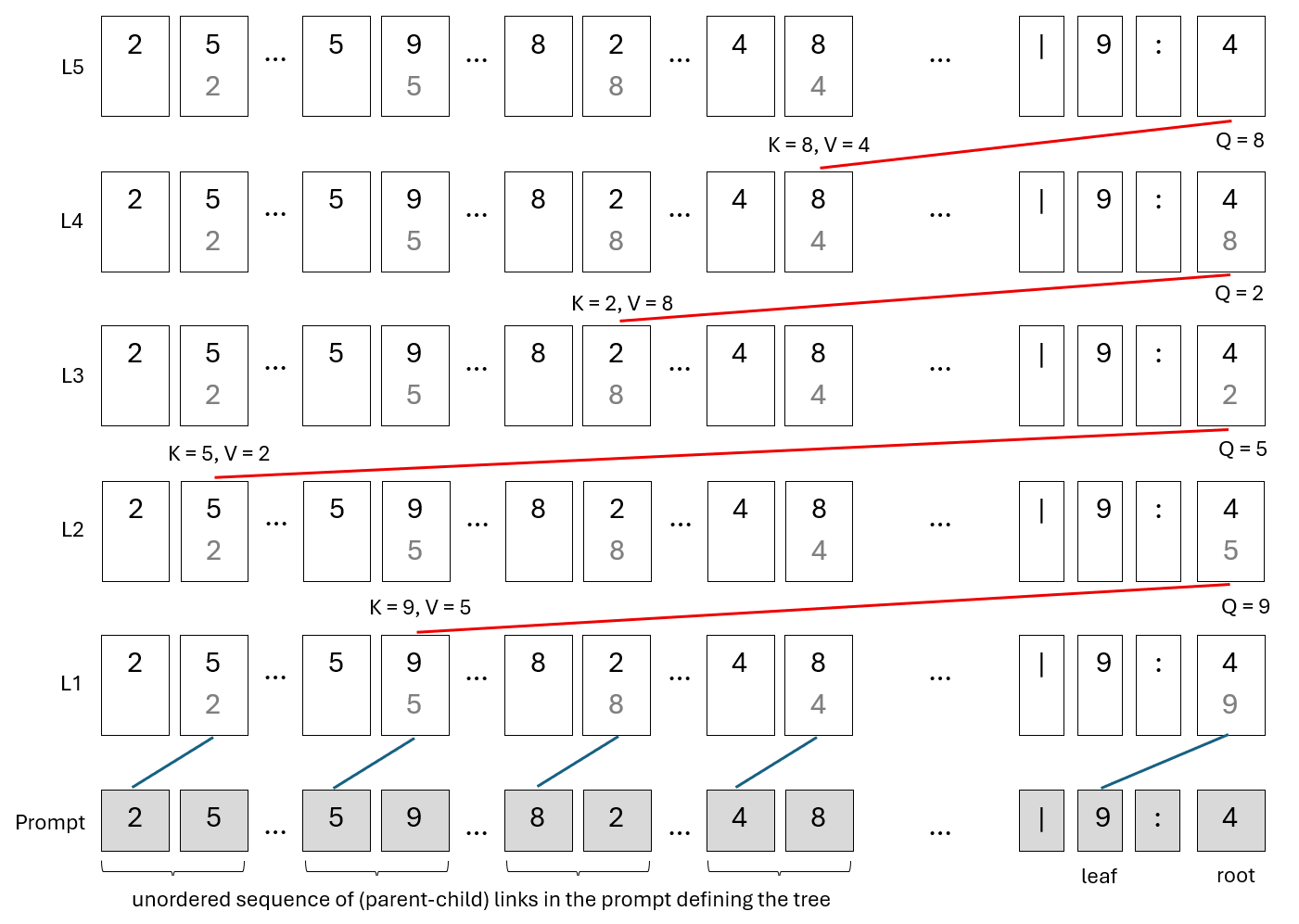}
\caption{Considering the binary tree example of Figure \ref{figure:TreeExample}, the root-to-leaf path to be generated is 4 8 2 5 9, where 4 is the root node and 9 is the leaf node. For simplicity, we use a model consisting of exactly 5 layers, matching the length of the path. The model searches for the backward path from the leaf 9 to the root 4, performing the computation at the position corresponding to the last token of the prompt, and storing in the corresponding residual stream the intermediate nodes it finds. Layer 1 copies the leaf 9 into the residual stream of the last position and, for each definition link of the tree, copies forward the parent node of the links; therefore, the prompt links of the form x → y are collapsed into single tokens starting from layer 2 (the copying heads are shown in blue in the figure). At layer 2, an induction head (in red) searches for a link x → 9, finds 4 → 9, and stores 4 into the residual stream. The same mechanism is applied in the subsequent layers until the root is reached. The backward path is now available in the residual stream of the last column and can be generated forward (one token at a time) starting from the top.}
\label{figure:VertReasoninExample}
\end{figure*}

As discussed in Section \ref{sec:task2}, the forward (root to leaf) navigation of a binary tree cannot be performed by deterministically following a single chain of links, because of the binary choices required at some intermediate nodes. Nevertheless, instead of implementing a full search approach (such as BFS or DFS) a simple algorithmic solution is moving backward from the given leaf to the root and then providing the node list in reverse order. This is exactly the approach implicitly learned by the model, as pointed out by the mechanistic explanation in \citep{Brinkmann2024}. It is crucial to note that the model has to compute the entire backward path before producing the first node of the output. In fact, the root node itself can have two children, and selecting the correct one requires knowing which branch leads to the desired leaf node. The vertical reasoning behind the learned approach is illustrated in Figure \ref{figure:VertReasoninExample}. We observe that the role of induction heads is here too central to unfold the reasoning chain along the vertical dimensions (layers), which takes place at the position of the last token in the prompt. 
Two other relevant findings of \citep{Brinkmann2024} are:
\begin{itemize}
    \item even if the model cannot exploit autoregression to verbalize intermediate reasoning steps, it uses positions associated to \textit{register} tokens in the prompt. Register tokens, acting as a sort of working memory, are often associated with tokens that do not contain useful information (e.g. punctuation) or tokens whose information has been copied to other positions and thus contain redundant information.
    \item when the path length (whose max is 15) exceeds the model depth (6 in the experiments), the illustrated backward chaining at the last token position is insufficient, and multiple partial paths are computed and merged in the last layers by the model. Even if the exact circuits and mechanism of this path merging are not explained in \citep{Brinkmann2024}, the authors provide some convincing evidence that these sub-paths are computed in parallel and stored in the register tokens.     
\end{itemize}

The second point is particularly relevant and coherent with other studies. For example, authors of \citep{Sanford2024} demonstrated that some classes of problems requiring $k$ steps, can be solved by a transformer of depth $O(log_2 k)$, exploiting internal parallel computation. The authors validate their theoretical findings on the \textit{k-hop induction heads} task (with $k$ up to 16) that, even if simpler than Task 2, still requires $k$ ops. Returning to root-to-leaf navigation, a well known result from graph theory is that given an adjacency matrix $A$ (i.e., where $A_{ij} = 1$ if there is an edge between nodes $i$ and $j$, 0 otherwise) the full reachability matrix $R$ (where $R_{ij} = 1$ if there is a path connecting node $i$ to node $j$) can be obtained by repeated squaring, where $(A + I)$ is iteratively squared with boolean products \citep{Rosen2012}. It can be easily proved that after the first squaring the reachability matrix includes all paths of max length = 2 and each successive squaring doubles the path max length, again leading to the above logarithmic rule. Therefore, in principle, if each transformer layer were able to perform a connection propagation step doubling the max path length, 6 layers would be enough to navigate a binary tree of depth 15 (being $log_2 15 < 6$). Of course, given the reachability matrix, the root-to-leaf navigation becomes straightforward.

After showing that a 6-layer transformer can successfully solve Task 2 through vertical reasoning, we now investigate whether a 2-layer model with no MLP (identical to the one used in Section \ref{sec:HorReas}) can achieve comparable performance relying on horizontal reasoning steered by the supplied CoT supervision. As expected, because a simple forward navigation cannot be implemented, validation accuracy drops substantially, reaching only 54\% after 50 epochs. Conversely, if we modify the task to perform leaf-to-root navigation, horizontal reasoning can make effective use of CoT supervision, and even a simple 2-layer transformer attains 99.7\% validation accuracy after only 2 epochs; moreover, we can observe the formation of induction-head-based circuits akin to those found for rule completion in Section \ref{sec:rulecompl}.

\section{Training without CoT}\label{sec:NoCot}

\subsection {Task 1}
When the model was trained as a binary classifier (i.e., the output is a single token 0 or 1), we were not able to reach convergence even using the full architecture ($l=6, h=8, d_{model}=128, MLP_{presence}=True$) and significantly increasing the training set size (from 4K to 150K examples). What we observed was a near-perfect accuracy on the training set (close to 100\%) with no generalization on the validation data (accuracy close to a random guess), indicating memorization rather than rule discovery. 
In principle, a vertical reasoning mechanism similar to that described in Section \ref{sec:Vert} and Figure \ref{figure:VertReasoninExample} could be exploited to unfold the reasoning chain along the layers in the position of the last token (with max 5 ops and 6 layers this would be possible even without internal parallel computation). However, it seems that simple binary supervision is not enough to induce such behavior. 
It is worth mentioning an additional experiment conducted during the course of this study. At that stage, the example generation process was different: all positive examples were full-length (5-ops), and an extra chain-breaking literal was used only for negative examples. Under this setup, the behavior of a model trained with CoT was very similar to that discussed in Section \ref{sec:MechExpl}. However, with some surprise, we were able to successfully train a simple model (2 layers, 2 heads, and MLP) without CoT supervision on a training set of 36864 examples. By analyzing its inference behavior, we found no evidence of induction heads. Instead, ad-hoc tests revealed that the model had learned to count the number of distinct literals in the prompt (a unit larger for negative examples). In other words, rather than learning the inference rules, the model exploited a bias in the training data to perform a simpler discrimination.

\subsection {Task 2}\label{subsec:Task2NoCot}
In the initial experiments we were able to achieve a reasonable convergence (in the range of 80-90\% after 50 epochs) with the full model on Task 2 even without CoT. This was quite surprising for us, because Task 2 without CoT can be seen as a generalization of Task 1 without CoT. A deep analysis of the intermediate results and comparison with training runs on Task 1 revealed that convergence was possible because of the generation bias (see Appendix \ref{sec:AppendixD}) that was leveraged by the model to solve the problem through some tricks, instead of really navigating the links. 
Once the bias was removed, as for Task 1, convergence was no longer achieved (around 50-60\% after 50 epochs). Here as well, we note that the reasoning strategy discussed in Section \ref{sec:Vert} could, in principle, be applied to solve the task; however, the absence of robust supervision failed to elicit this behavior.

\section{CoT vs Curriculum Learning}

\begin{figure}[ht]
\centering
\includegraphics[scale=0.35]{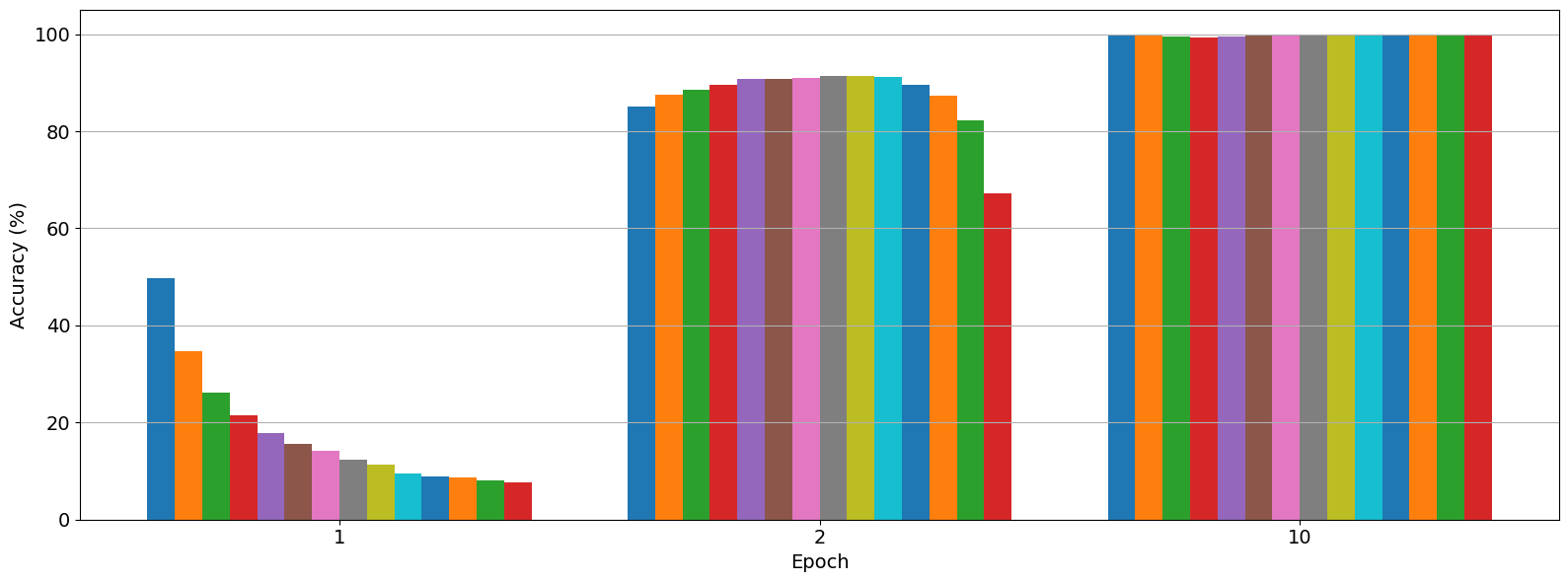}
\caption{The three groups of bars represent token accuracy (on the validation set) after epochs 1, 2, and 10. The 15 bars within each group correspond to the accuracy for tokens at distances 1 through 15 from the target leaf, respectively. Accuracy here is not computed autoregressively; instead, as during training, the model can see the ground-truth tokens for all preceding positions. After epoch 1, accuracy is substantially higher for nodes located closer to the target leaf node, and as training progresses, accuracy tends to level out across the different distances. }
\label{figure:CL1}
\end{figure}

\begin{figure}[ht]
\centering
\includegraphics[scale=0.35]{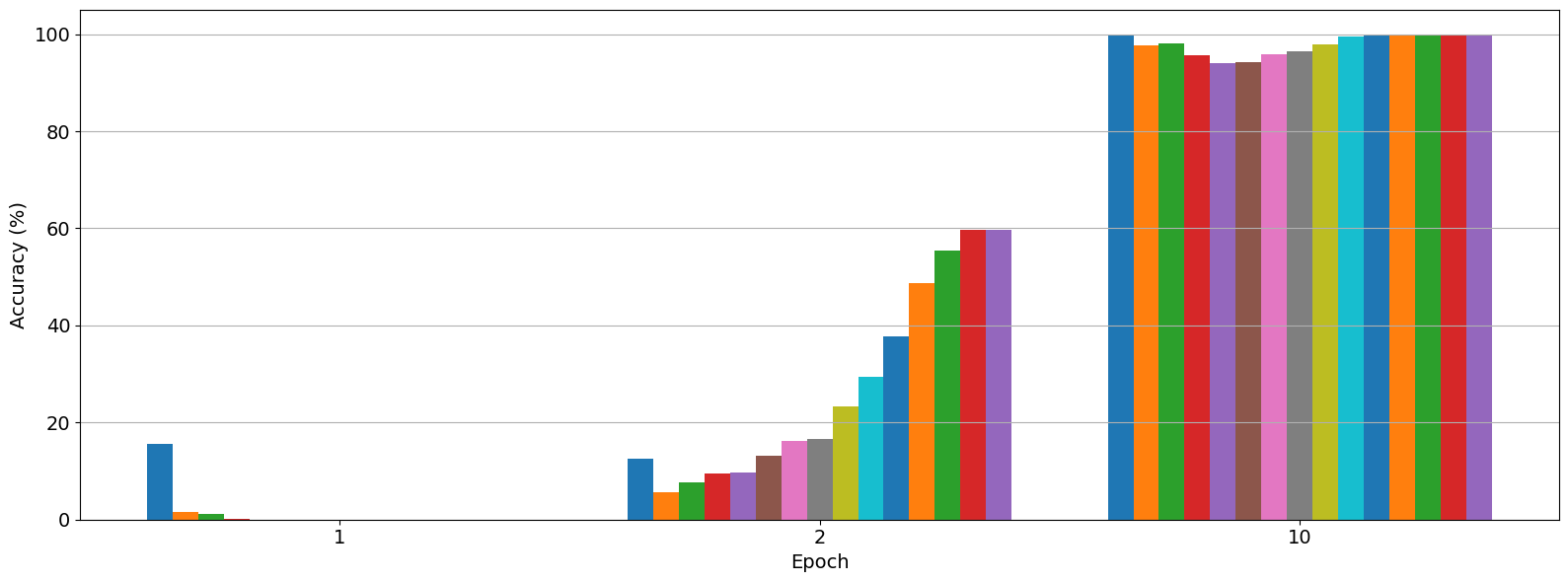}
\caption{The three groups of bars represent the sequence accuracy on the validation set after epochs 1, 2, and 10. The 15 bars within each group correspond to the accuracy for examples of different lengths (from 1 to 15 Ops). Accuracy is computed in autoregressive manner. After epoch 1, the model correctly solves only some examples with one or very few operations, whereas by epoch 10 it can reliably handle more complex examples. Notably, at epoch 2, examples of intermediate length are the most challenging for the model; this is likely because their associated trees are more balanced and thus require more binary decisions compared to filiform trees of maximum depth. }
\label{figure:CL2}
\end{figure}

Now, focusing on Task 2, an interesting question is why the model without CoT does not converge (\ref{subsec:Task2NoCot}) while, using CoT for the root-to-leaf navigation (\ref{sec:Vert}), it converges. After all, in the root-to-leaf navigation, generating the first token can require solving the reachability problem for a far leaf, without helpful information from CoT on how such a problem could be solved, thus making the task somewhat similar to learning without CoT.
To further investigate this issue, we ran additional experiments, by annotating the details of the examples correctly processed at the different training stages. What we observed is that the CoT provided, even if not able to provide step-by-step details on how to solve the forward tree navigation, induces a sort of Curriculum Learning where the model can initially learn simplified patterns and then leverage the learned rules to solve more complex cases. Such curriculum learning can take place along two different dimensions:

\begin{itemize}
\item the first tokens in the output are difficult to predict because require to solve a far-reachability problem, while the last tokens are simpler to predict because the corresponding nodes are closer to the target leaf. Experiments reveal that during initial iterations, the last tokens in the output are predicted with much higher accuracy (see Figure \ref{figure:CL1}).
\item examples have different root-to-leaf path lengths. Experiments reveal that the shorter examples (requiring less Ops) are more likely to be solved since from the first training iterations, while the longer ones need more epochs (see Figure \ref{figure:CL2})
\end{itemize}

So our results indicate that in the considered logic deduction tasks, CoT can bring advantages to both horizontal and vertical reasoning: in the former it guides the model to learn the solution by decomposing step-by-step the problem, in the latter it can enable a sort of curriculum learning, allowing implicit learning to first address simpler problems and then reuse the knowledge acquired for the more difficult ones.

\section{Conclusions}

In this paper, we investigated how small transformer-based language models can learn to solve multi-step deductive reasoning tasks. We considered two symbolic tasks with substantially different reasoning requirements: a logical consequence task over chains of implications, requiring up to five inference steps, and a root-to-leaf navigation task in binary trees, requiring paths of up to fifteen steps. Despite the simplicity of the input language, both tasks have large generation spaces and cannot be solved by mere memorization when training and validation sets are properly separated.

Our results show that small LMs trained from scratch with CoT supervision can learn the underlying inference rules and develop internal mechanisms that implement structured deductive reasoning. This finding contrasts with our previous study on arithmetic reasoning \citep{Maltoni2024}, where the analyzed model appeared to encode tokens into a value space and behave more like a statistical regressor. Taken together, these results support the view that language models may rely on different computational strategies depending on the nature of the task, ranging from symbolic rule-based mechanisms to more sub-symbolic numerical approximations.

For the logical consequence task, we provided a detailed mechanistic explanation of the horizontal reasoning process learned by a shallow attention-only model. In this setting, CoT supervision enables the model to decompose the task into explicit intermediate steps, which are generated autoregressively and then reused to continue the inference chain. The resulting computation is highly interpretable: rule completion, rule chaining, and final decision making can be associated with identifiable attention-based circuits. In particular, induction-head-like mechanisms play a central role by searching for matching rule antecedents and retrieving the corresponding consequents. The analysis was further supported by residual-stream decoding and by the proposed truncated-pseudoinverse technique for inspecting the information carried by queries, keys, and values.

For the binary-tree navigation task, we examined a different form of reasoning, which we referred to as vertical or implicit reasoning. In this case, the model cannot simply rely on previously generated tokens to determine the first output token; rather, a substantial part of the computation must be performed internally across the layer hierarchy before generation begins. Our experiments suggest that CoT supervision still plays an important role in this setting, but in a different way: instead of directly providing a step-by-step horizontal decomposition, it appears to induce a form of curriculum learning. The model first learns simpler sub-problems, such as shorter paths or later output positions that are closer to the target leaf, and progressively extends this competence to more complex cases.

The experiments without CoT further clarify the importance of the supervision signal. When models are trained only with binary outputs, learning the intended deductive procedure becomes substantially harder. In the absence of exploitable dataset biases, the models tend either not to generalize or to memorize the training data. Conversely, when biases are present in the generation process, models may exploit them to obtain good performance without learning the intended logical algorithm. This reinforces the need for careful dataset design and for mechanistic analysis when evaluating reasoning abilities in language models.

The study also has limitations. The low-level analysis of the first task benefits from the use of shallow models, shared input/output embeddings, and simplified symbolic syntax; these conditions make techniques such as LogitLens and truncated-pseudoinverse decoding particularly effective, but they may not directly transfer to larger pretrained models or more naturalistic settings. For the second task, although the emergence of vertical reasoning is evident, the mechanisms used to merge partial paths when the reasoning depth exceeds the number of layers are not yet fully understood.

Several directions remain open for future work. More complex logical tasks, including cases requiring backtracking or multiple alternative proof paths, could help assess whether the mechanisms identified here scale beyond chain-like inference. For tree navigation, a deeper investigation of path fusion could clarify its relationship with parallel computation and repeated-squaring-like strategies for reachability. Finally, our results highlight the broader relevance of induction heads beyond text continuation and in-context learning. In our experiments, they constitute a fundamental component of circuits implementing logical inference. A particularly interesting direction is to explore whether attention mechanisms explicitly decoupling key and value positions could support induction-head-like behavior even in one-layer circuits.

\vspace{6pt} 

\appendixtitles{yes} 
\appendixstart
\appendix
\section[\appendixname~\thesection]{Generation of the synthetic dataset for Task 1} \label{sec:AppendixA}

The generation process uses a vocabulary of $n$ literals ($n=20$ in our experiments) and produces chains of $m$ implications ($m=5$ in our experiments). The generated dataset has an equal number of positive and negative examples. To ensure that positive examples closely resemble negative ones, we first generate a negative example and, if necessary, apply a straightforward modification (described below) to convert it into a positive one. The generation of a negative example proceeds through the following steps:

\begin{enumerate}
    \item Random selection of $m+2$ distinct literals (without repetition and with order) from the set of $n$ literals.
    \item The first $m+1$ literals $[l_1,l_2,l_3,...,l_{m+1}]$ are used to form $m$ implications: $[l_1 \rightarrow l_2,l_2 \rightarrow l_3,...,l_m \rightarrow l_{m+1}]$.
    \item The head $q_0$ and tail $q_1$ of the query implication are set to $l_1$ and $l_{m+1}$, respectively.
    \item A break point $b \in [1,...,m]$ is randomly selected. The implication $l_b \rightarrow l_{b+1}$ is modified to break the chain by injecting the extra literal $l_{m+2}$. A coin toss determines whether to replace the head ($l_b$) or the tail ($l_{b+1}$).
    \item The resulting list of $m$ implications is randomly shuffled.
\end{enumerate}

The number of different negative examples is:

\begin{center}
$P^n_{m+2} \cdot m \cdot 2 \cdot m!$
\end{center}

\noindent
where $P^n_{m+2}$ denotes the number of permutations of $m+2$ elements chosen from $n$ (Step 1), $m \cdot 2$ refers to Step 4, and $m!$ corresponds to the final shuffling (Step 5). For $n=20$ and $m=5$, this yields 468,840,960,000 possible generations.

A negative example with $b>1$ (we need at least one valid implication in the chain) can be deterministically converted into a positive example by setting $q_1=l_{m+2}$.

Although the probability of duplication is extremely low given the huge generation space, a hash table was used during dataset creation to ensure that the same sequence is not generated more than once, thereby preventing overlap between the training and validation sets.

\section[\appendixname~\thesection]{Truncated pseudoinverse for query, key, and value decoding} \label{sec:AppendixB}

To analyze the information extracted by the Query, Key, and Value matrices in the transformer model, we employ a truncated pseudoinverse technique based on Singular Value Decomposition (SVD). Our approach projects vectors from their subspace back into the original residual stream space, facilitating interpretability through token-level decoding.

When operating with a single head ($h=1$), the projection matrices $W_Q$, $W_K$, and $W_V$ are squared matrices in $\mathbb{R}^{d_{model} \times d_{model}}$, typically full rank. These matrices extract query, key, and value information from the residual stream by projection into subspaces. Since a projection through a full-rank matrix does not reduce dimensionality, all the information, even if rearranged, is preserved (i.e., retro-projection using the exact inverse would recover the original residual stream content). Therefore, to focus on the most relevant information extracted by the Query, Key, and Value matrices, we employ a truncated pseudoinverse approach. 

Let us consider a generic matrix $W \in \mathbb{R}^{d_{model} \times d_{model}}$ (the same process applies to $W_Q$, $W_K$, and $W_V$). Its SVD decomposition is given by:

\begin{center}
$W=U \Sigma V^T$
\end{center}

\noindent
where $U$ and $V$ are orthogonal matrices, and $\Sigma$ is a diagonal matrix containing the singular values $\sigma_1,\sigma_2,...,\sigma_{d_{model}}$, sorted in descending order ($\sigma_1 \ge \sigma_2 \ge ... \ge \sigma_{d_{model}}$). To emphasize the directions in the subspace that carry the most significant information, we retain only the top $k$ singular values, where $k$ is selected such that the cumulative sum:

\begin{center}
$s_k=\frac{\sum^k_1 \sigma_i}{\sum^{d_{model}}_1 \sigma_i}$
\end{center}

\noindent
reaches a desired threshold (e.g., 95\%). The truncated pseudoinverse is then defined as:

\begin{center}
$W_k^+=V_k \Sigma_k^+ U_k^T$
\end{center}

\noindent
where $V_k$, $\Sigma_k$, and $U_k$ are the matrices corresponding to the $k$ top singular values, and $\Sigma_k^+$ contains their reciprocals.

In our experiments, we noted that: (i) for $W_Q$ and $W_V$ the optimal threshold $s_k$ lies in the range $[0.75,0.85]$, resulting in a dimensionality reduction from 128 to around 50; (ii) for $W_K$, a higher threshold $s_k$ is required (i.e., $[0.95,0.99]$), leading to a reduction from 128 to around 110 dimensions. Further work is necessary to determine an optimal (automatic) tuning of these thresholds for interpretability studies.

When operating with multiple heads, the projection matrices of single attention heads are no longer square, and their projection and retro-projection (by pseudoinverse) already involve some information loss. In this scenario, the use of a truncated pseudoinverse may still be beneficial for discarding additional information.

\section[\appendixname~\thesection]{Models and training details} \label{sec:AppendixC}

Table \ref{table:LMDetails} summarizes the details of the NanoGPT models used in Task 1 and Task 2 experiments (see Sections \ref{sec:HorReas} and \ref{sec:Vert}, respectively).

\begin{table}[ht]
\caption{Details of the NanoGPT models used in Task 1 (Section \ref{sec:HorReas}) and Task 2 (Section \ref{sec:Vert}) experiments.}
\label{table:LMDetails}

\scriptsize
\renewcommand{\arraystretch}{1.15}
\setlength{\tabcolsep}{3pt}

\begin{tabularx}{\linewidth}{
    @{}
    >{\raggedright\arraybackslash}p{2.7cm}
    >{\raggedright\arraybackslash}X
    >{\raggedright\arraybackslash}X
    @{}
}
\toprule
& \textbf{Task 1} & \textbf{Task 2} \\
\midrule

Tokenization & Single characters are used as tokens & Numbers from 0 to 15 are used as tokens \\
\midrule

Vocabulary size & 28 & 20 \\
\midrule

Vocabulary 
& {[$'A'$,\allowbreak ...,\allowbreak $'T'$],\allowbreak $'@'$,\allowbreak $'|'$,\allowbreak $','$,\allowbreak $'>'$,\allowbreak $'\_'$,\allowbreak $'-'$,\allowbreak $'0'$,\allowbreak $'1'$} 
& {[$'0'$,\allowbreak ...,\allowbreak $'15'$],\allowbreak $'@'$,\allowbreak $'|'$,\allowbreak $','$,\allowbreak $':'$} \\
\midrule

Prompt length, output length & 24, 23 & 49, variable in the range $[1,...,15]$ \\
\midrule

Token embedding & Learned (Token encoding and decoding share weights) & Learned \\
\midrule

Positional encoding & Learned & Learned \\
\midrule

$d_{model}$ & 128 & 128 \\
\midrule

$d_{ff}$ & - & $d_{model} \times 4$ \\
\midrule

\textit{h} & 1 & 1 \\
\midrule

\textit{l} & 2 & 6 \\
\midrule

$MLP_{presence}$ & False & True \\
\midrule

Learnable parameters & 144,384 & 1,203,605 \\

\bottomrule
\end{tabularx}
\end{table}

\begin{figure}[ht]
\centering
\includegraphics[scale=0.7]{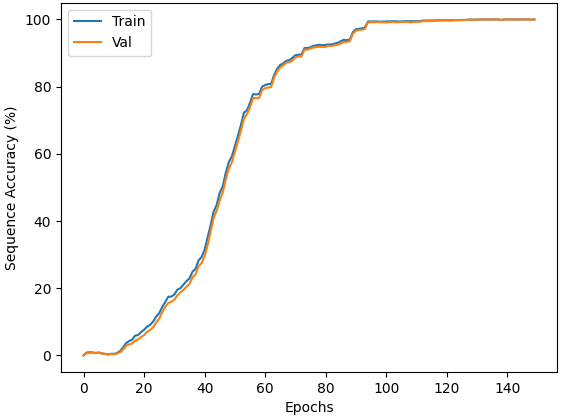}
\caption{Average accuracy over 20 runs reaching convergence on Task 1.}
\label{figure:AvgTrainGraphTask1}
\end{figure}

\begin{figure}[ht]
\centering
\includegraphics[scale=0.85]{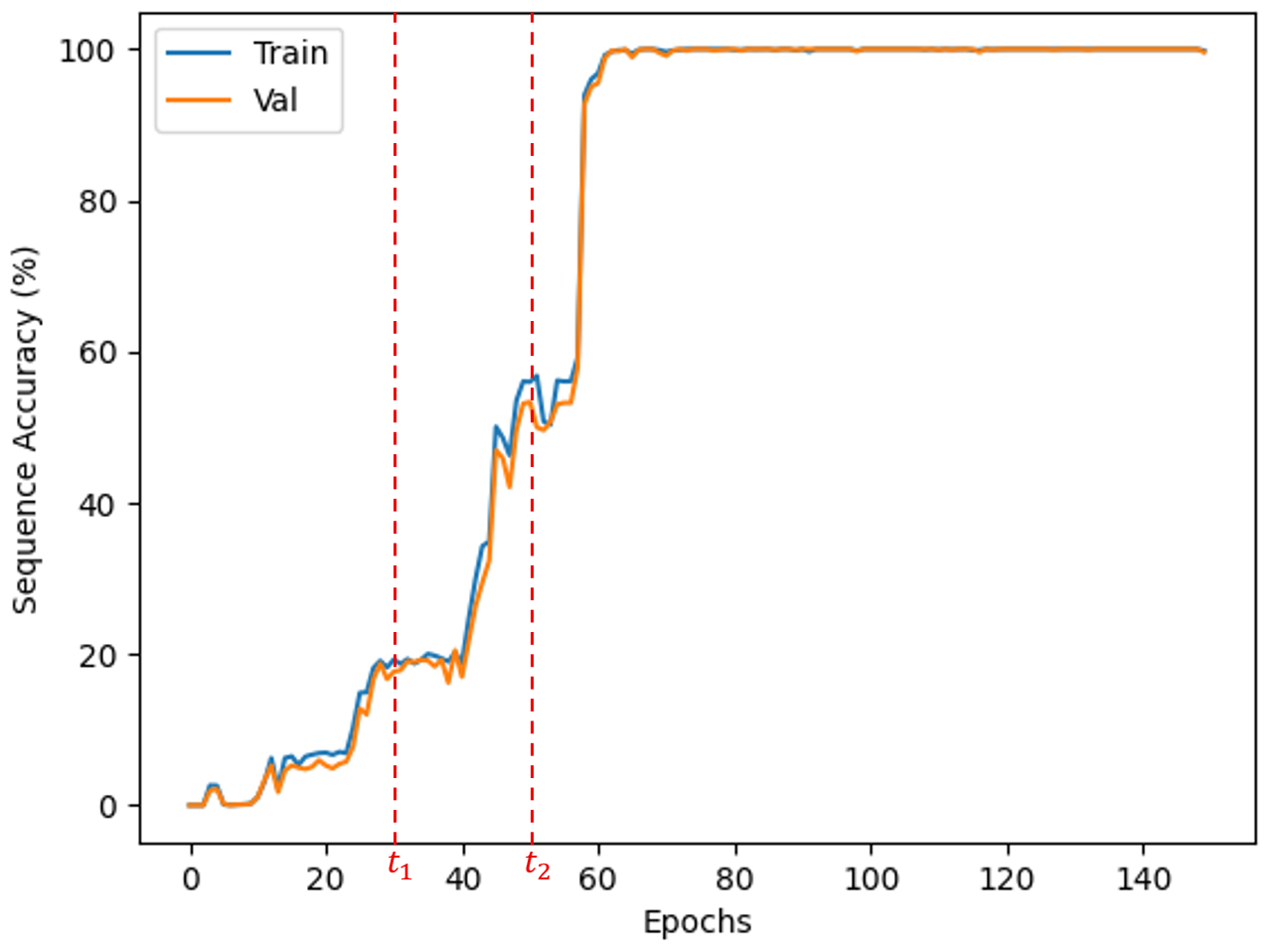}
\caption{The model convergence during a specific run on Task 1, with identification of points of interest $t_1$ and $t_2$.}
\label{figure:SpecificTrainingGraphWithPointOfInterest}
\end{figure}

Despite the small size of the Task 1 training set (compared to the vast input space), the LM learns the task (accuracy is \textasciitilde100\%) in just a few hundred epochs, as shown in Figure \ref{figure:AvgTrainGraphTask1}. It also generalizes perfectly on the validation set. When repeating the experiment with different training sets and different random initial weights for the LM, we observed that approximately 20\% of the runs converged within 250 epochs. The convergence rate increased to 80\% when using the full NanoGPT architecture. However, for the purposes of this study, the success rate across runs is not critical — our goal is to analyze the inference behavior of a few models that have reached convergence.
Figure \ref{figure:SpecificTrainingGraphWithPointOfInterest} shows the convergence of the model during a specific training run (the same one used in Section \ref{sec:HorReas}). Two points of interest can be observed where the model accuracy, initially close to random guessing, suddenly increases. At $t_1$ (epoch 30), accuracy reaches 20\%; at $t_2$ (epoch 50), it further grows to 60\%, with final convergence occurring around epoch 65\footnote{A similar convergence trajectory was observed in other runs, with two intermediate plateaus at approximately 20-30\% and 60-80\% accuracy before reaching final convergence.}. By inspecting the checkpoints at $t_1$ and $t_2$ using the interpretability techniques discussed in Section \ref{sec:HorReas} and computing additional statistics, we discovered that:

\begin{itemize}
    \item At $t_1$, the model has already established an initial implementation of induction heads responsible for rule completion and rule chaining, but they work well only on the initial implications in the chain. We guess that some aspects related to positional encoding have not yet been fully resolved. Such focus on initial implications aligns with the distribution of the training data, where most examples contain one or a few correct initial implications.
    \item At $t_2$, the rule-chaining and rule-completion mechanisms are fully working. In fact, sequence accuracy computed by excluding the last token (i.e., the final decision) has already reached 100\%. However, the final decision circuit explained in Subsection \ref{subsec:StartFinalDec} (particularly attention links 61-65 in Figure \ref{figure:CircuitsStartFinalDecision}) has not yet formed. Consequently, the token accuracy of the last token (54.7\%) is just slightly higher than a random guess.
\end{itemize}

Consistent with previous findings \citep{Nanda2023}, we observed that the emergence of induction heads leads to a sudden improvement in generalization. However, our results also suggest that a multi-stage improvement could be necessary to refine the underlying mechanisms.

\begin{figure}[ht]
\centering
\includegraphics[scale=0.7]{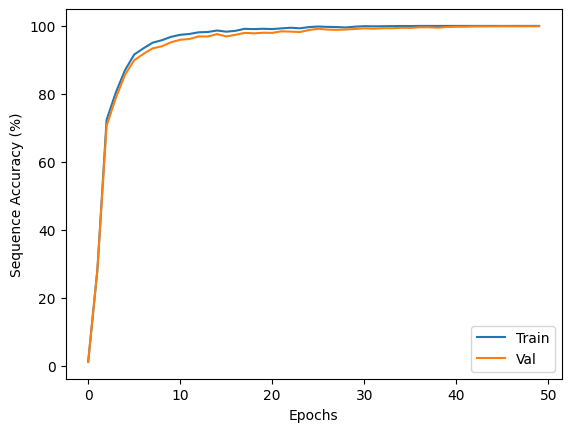}
\caption{Accuracy on Task 2.}
\label{figure:TrainGraphTask2}
\end{figure}

Figure \ref{figure:TrainGraphTask2} shows the convergence trend of a typical run in Task 2. It is worth noting that the training set here is about 36 times larger than that used in Task 1, so the number of iterations (i.e., mini-batches) is substantially greater.

\section[\appendixname~\thesection]{Original generation bias in Task 2} \label{sec:AppendixD}
\begin{figure}[H]
\centering
\includegraphics[scale=0.8]{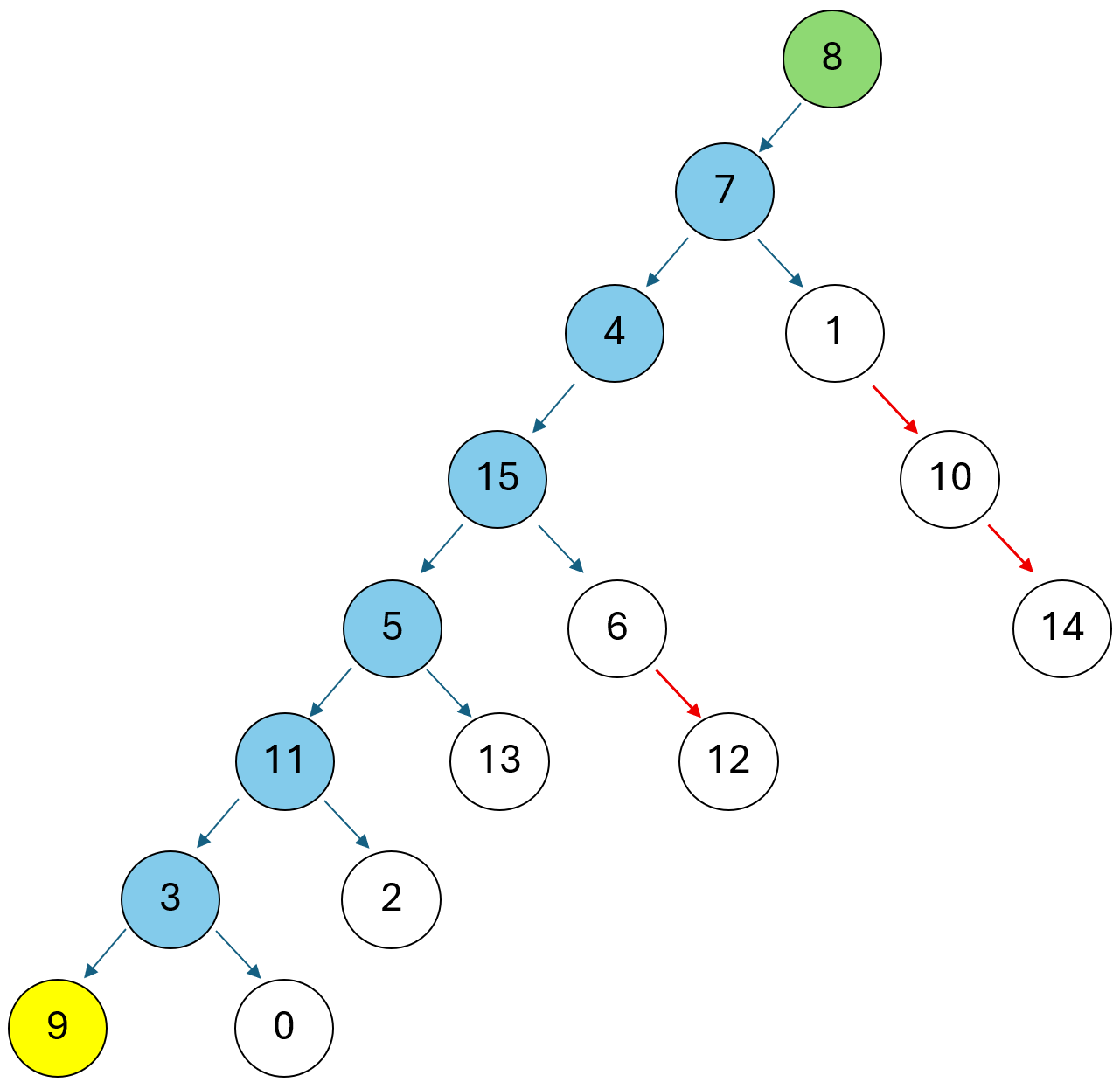}
\caption{An example of a biased binary tree: the root (green), the target leaf (yellow), the main path nodes (light blue), and the biased parent-child pairs (red arrows).}
\label{figure:BiasesTreeExample}
\end{figure}

For the Task 2 experiment discussed in Section \ref{sec:Vert}, we employed the first 150K examples from the file “dataset.txt” provided by the authors of \citep{Brinkmann2024} in the repository \citep{Brinkmann2024git}.
All 150K binary trees were produced with the \textit{generate\_example} function contained in the file “gen.py”.
The function first constructs the main path with a fixed length by randomly selecting its nodes. Then, the remaining nodes are attached to the tree at random locations, ensuring that the previously created main path is not further extended. In this second phase, however, node insertion follows an ordered scheme: nodes with lower IDs are placed before those with higher IDs.
This procedure introduces a bias that a model can leverage to solve the task even in the absence of CoT. Specifically, for any parent–child pair where the parent is not on the main path, the child node ID is always larger than the parent node ID (see pairs 1→10, 10→14, and 6→12 in Figure \ref{figure:BiasesTreeExample}).

\reftitle{References}

\end{document}